\documentclass{article} % For LaTeX2e
\usepackage{nips13submit_e,times}
\usepackage{hyperref}
\usepackage{url}

\usepackage{times}

\usepackage{latexsym}

\usepackage{amsmath}

\usepackage{amssymb}

\usepackage{multirow}

\usepackage{tikz}

\PassOptionsToPackage{dvips}{graphicx}

\usetikzlibrary{fit,positioning}

\usepackage{booktabs}

\usepackage{yates}
\usepackage{caption}
\usepackage{subcaption}
\usepackage{verbatim}

\newcommand{\baseline}{{\sc Baseline}}
\newcommand{\scl}{{\sc SCL}}
\newcommand{\hmm}{{\sc HMM-80}}
\newcommand{\huanghmm}{{\sc HuangHMM-80}}

\newcommand{\brown}{{\sc Brown-1000}}
\newcommand{\embedding}{{\sc Embeddings-50}}

\newcommand{\fhmmr}[2]{{\sc FHMM-#1-#2}}
\newcommand{\fhmmpostr}[2]{{\sc FHMM-POST-#1-#2}}

\DeclareMathOperator*{\argmax}{arg\,max}
\DeclareMathOperator*{\argmin}{arg\,min}

\title{Factorial Hidden Markov Models for Learning Representations of Natural Language}

\author{
Anjan Nepal \\
Temple University\\
Broad St. and Montgomery Ave.\\
Philadelphia, PA 19122\\
\texttt{anjan.nepal@temple.edu} \\
\And
Alexander Yates \\
Temple University\\
Broad St. and Montgomery Ave.\\
Philadelphia, PA 19122\\
\texttt{yates@temple.edu} \\
}

\date{}

\nipsfinalcopy % Uncomment for camera-ready version

\begin{document}

\maketitle

\begin{abstract}
Most representation learning algorithms for language and image processing are local, in that they identify features for a data point based on surrounding points.  Yet in language processing, the correct meaning of a word often depends on its global context.  As a step toward incorporating global context into representation learning, we develop a representation learning algorithm that incorporates joint prediction into its technique for producing features for a word.  We develop efficient variational methods for learning Factorial Hidden Markov Models from large texts, and use variational distributions to produce features for each word that are sensitive to the entire input sequence, not just to a local context window.  Experiments on part-of-speech tagging and chunking indicate that the features are competitive with or better than existing state-of-the-art representation learning methods.
\end{abstract}

\section{Introduction}

Most existing representation learning algorithms project and transform the local context of data to produce their representations.  Yet in many applications, particularly in natural language processing (NLP), joint prediction is crucial to the success of the application, and this requires reasoning over global structure.

We investigate a representation learning approach based on Factorial Hidden Markov Models \cite{factorial-hmms}.  Like spectral methods \cite{dhillon_icml12_tscca} and neural network methods \cite{semi-supervised-embedding-srl,2008-icml-collobert-weston} for learning representations in NLP, FHMMs provide a \emph{distributed} representation of words, meaning that they assign a multi-dimensional vector of features to each word, although in our models each latent dimension is discrete rather than real-valued.  The factored latent states in the FHMMs make it tractable to train models with exponentially larger latent state spaces than those permitted by regular Hidden Markov Models (HMMs) \cite{2014-compling-huang-lvlm-rep-learning,smoothing-HMM} or Brown clusters \cite{1992-brown-clustering,2010-acl-word-representations-turian}, and empirically, models with larger latent state spaces produce better language models and better estimates of latent features for information extraction \cite{2010-naacl-ddowney-perplexity-vs-accuracy}.  More importantly, FHMMs provide the ability to assign a representation to a word that is sensitive to the full observation sequence.  In contrast, both spectral methods and neural network models rely on fixed-length context windows to produce word representations.

Our contributions in this paper are mainly two fold: 1) We develop novel variational approximation algorithms for FHMMs with discrete-valued variables and 2) we provide empirical evaluation of the distributed representations from FHMMs which can provide different representations per word depending on the whole sentence context. We compare the performance against the systems that produce either only global context dependent but not distributed representations, or fixed representations per word which are not distributed, or fixed distributed representations per word which are dependent on the local context of the sequence.

%Learning and inference for FHMMs can be computationally expensive.  We develop algorithms for FHMMs with discrete-valued variables that can tractably handle medium-sized text datasets on one machine, and that can in principle scale to larger datasets using distributed implementations.  In particular, we investigate variational inference and learning techniques that are suitable for discrete-valued FHMMs.  Our largest and best-performing model was trained to convergence over a dataset of nearly three million tokens on a single quad-core machine in under 10 days.

In experiments, we investigate the ability of FHMMs to provide useful representations for part-of-speech (PoS) tagging and noun-phrase chunking.  In comparison with existing HMM representations, we find that our FHMM-based representations are more scalable to large latent state spaces and provide better application performance, reducing error by 4\% on words that never appear in labeled PoS-tagging training data, and by 5\% on chunks whose initial word never appears in labeled chunking training data.

The next section discusses previous work in representation learning for NLP.  Section \ref{sec:fhmm} introduces our FHMM model for representation learning, and Section \ref{sec:learningandinference} presents the inference and learning algorithms.  Section \ref{sec:experiments} describes our empirical evaluation of the model as a technique for learning latent features of words for subsequent classification tasks.  Section \ref{sec:conclusion} concludes.

\section{Previous Work}
\label{sec:previous_work}

There is a long tradition of NLP research on representations, mostly falling into one of four categories:  1) vector space models of meaning based on document-level lexical cooccurrence statistics \cite{saltonIR,2010-jair-turney-pantel,2006-sahlgren}; 2) dimensionality reduction techniques for vector space models \cite{latent-semantic-indexing,1997-self-organizing-maps-as-reps,1998-kaski-dim-reduction-by-random-mapping,2005-random-indexing-for-word-reps,lda,2007-ica-for-semantic-features}; 3) using clusters that are induced from distributional similarity \cite{1992-brown-clustering,1993-acl-pereira-distributional-clustering,1998-trigram-brown-clustering} as non-sparse features \cite{2009-acl-ner-with-word-clusters,2009-parsing-with-brown-clusters,2008-koo-dep-parsing-with-word-clusters,zhao-conll09}; 4) and recently, language models \cite{2008-neural-net-lang-models,2009-nips-mnih-hierarchical-lang-model} as representations \cite{collobert:2011b,semi-supervised-embedding-srl,2008-icml-collobert-weston,2009-bengio-curriculum-learning}, some of which have already yielded state of the art performance on domain adaptation tasks \cite{2014-compling-huang-lvlm-rep-learning,2010-danlp-lvlms-for-domain-adaptation,huang-open-domain-srl,2010-acl-word-representations-turian,smoothing-HMM} and information extraction tasks \cite{2010-naacl-ddowney-perplexity-vs-accuracy,downey07realm}.  Our work combines the power of distributed representations from the neural network models with the joint inference from the HMM-based clustering approaches.

To the best of our knowledge, FHMMs have not been used for learning representations in NLP or elsewhere, and they are rarely used with discrete-valued observations.  This may be in part due to the perception that training with discrete-valued observations is too computationally challenging, at least for real NLP datasets.  Jacobs \etal\ \cite{factorial-hmms-backfitting} use a generalized backfitting approach for training a discrete-observation FHMM, but they have not run experiments on a naturally-occurring dataset, and they focus on language modeling rather than representation learning. We use a different training procedure based on variational EM \cite{variational-methods}, and provide empirical results for a representation learning task on a standard dataset.

Recent work by Socher \etal\ \cite{socher-acl-2013} has annotated parse trees with latent vectors that compose hierarchically.  Like our model, these techniques operate on distributed representations for natural language, but as with neural network models they do not perform any kind of joint prediction over the whole structure to optimize the representations for individual words. Titov and Henderson \cite{2007-acl-titov-sigmoid-belief-net}\ and Henderson and Titov \cite{2010-jmlr-henderson-sigmoid-belief-net}\ also learn representations that are affected by global context and also use variational approximation for inference but apply the technique to parsing. Grave \etal\ \cite{grave2013hidden}\ use mainly two types of latent variable models: one essentially an HMM and the other has the latent variables associated in a syntactic dependency tree. While they also get the latent states for the words that are both context dependent and globally affected, our method can handle exponentially large state space and the representations are distributed.

Most previous work on domain adaptation has focused on the case where some labeled data is available in both the source and target domains \cite{frustratingly-easy-domain-adaptation,instance-weighting,max-ent-domain-adaptation-three-model-combo,2009-naacl-finkel-domain-adaptation-ner,2010-machine-learning-multi-source-domain-adaptation,2008-emnlp-multi-source-adaptation}.  Learning bounds are known \cite{domain-adaptation-theory,2009-nips-multi-source-domain-adaptation-theory}.  Daum\'e III \etal\ \cite{2010-danlp-daume-semi-sup-domain-adaptation} use semi-supervised learning to incorporate labeled and unlabeled data from the target domain.  In contrast, we investigate a domain adaptation setting where no labeled data is available for the target domain.  %Solutions to this task can readily apply to new domains and large corpora like the Web with no manual effort.
%We seek techniques that require no new labeled examples for new domains, so that the system will approach the ideal of a train-once, use-anywhere system.  
%In lieu of {\em labeled} examples for new domains, our techniques leverage the large amounts of readily-available unlabeled text from new domains, in combination with labeled data from the training domain. This allows us to apply NLP systems effectively to domains with no pre-existing training data at no cost.  It also allows us to apply NLP systems to large, diverse corpora like the Web, where it is impractical to manually label examples for every domain.  

\section{The Factorial Hidden Markov Model for Learning Representations}
\label{sec:fhmm}

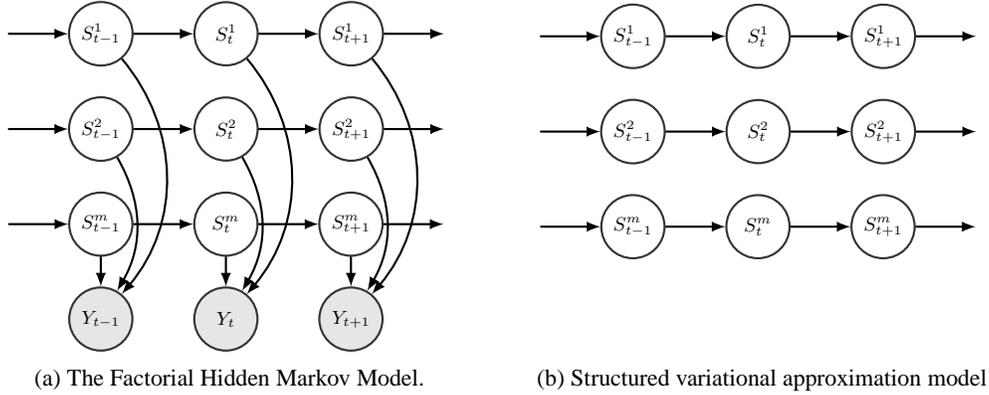
\begin{figure}[t]
	\centering
		\begin{subfigure}[b]{0.45\textwidth}
			\begin{tikzpicture}[scale=0.8, font=\small, every node/.append style={transform shape}]
\tikzset{dummy/.style={}}
\tikzstyle{var}=[circle, minimum size = 10.5mm, thick, draw =black!80, node distance = 5mm and 10mm]
\tikzstyle{connect}=[-latex, thick]
\tikzstyle{arc1}=[-latex, thick, bend left=40]
\tikzstyle{arc2}=[-latex, thick, bend left=30]
  \node[var] (S_t-1_1){$S_{t-1}^1$};
  \node[dummy, name=input1, left=of S_t-1_1] {};
  \node[var] (S_t_1) [right=of S_t-1_1] {$S_{t}^1$};
  \node[var] (S_t+1_1) [right=of S_t_1] {$S_{t+1}^1$};
  \node[dummy, name=output1, right=of S_t+1_1] {};
  \node[var] (S_t-1_2) [below=of S_t-1_1]{$S_{t-1}^2$};
  \node[dummy, name=input2, left=of S_t-1_2]{};
  \node[var] (S_t_2) [right=of S_t-1_2]{$S_{t}^2$};
  \node[var] (S_t+1_2) [right=of S_t_2]{$S_{t+1}^2$};
  \node[dummy, name=output2, right=of S_t+1_2] {};
  \node[var] (S_t-1_m) [below=of S_t-1_2]{$S_{t-1}^m$};
  \node[dummy, name=inputm, left=of S_t-1_m]{};
  \node[var] (S_t_m) [right=of S_t-1_m]{$S_{t}^m$};
  \node[var] (S_t+1_m) [right=of S_t_m]{$S_{t+1}^m$};
  \node[dummy, name=outputm, right=of S_t+1_m] {};
  
  \node[var, fill = black!10] (Y_t-1) [below=of S_t-1_m]{$Y_{t-1}$};
  \node[var, fill = black!10] (Y_t) [right=of Y_t-1]{$Y_{t}$};
  \node[var, fill = black!10] (Y_t+1) [right=of Y_t]{$Y_{t+1}$};
  
 \path (input1) edge [connect] (S_t-1_1)
 	(S_t-1_1) edge [connect] (S_t_1)
        (S_t_1) edge [connect] (S_t+1_1)
        (S_t+1_1) edge [connect] (output1)        
        (input2) edge [connect] (S_t-1_2)
	(S_t-1_2) edge [connect] (S_t_2)
        (S_t_2) edge [connect] (S_t+1_2)
	(S_t+1_2) edge [connect] (output2)		
	(inputm) edge [connect] (S_t-1_m)
        (S_t-1_m) edge [connect] (S_t_m)
        (S_t_m) edge [connect] (S_t+1_m)
        (S_t+1_m) edge [connect] (outputm)

	(S_t-1_1) edge [arc1] (Y_t-1)
	(S_t-1_2) edge [arc2] (Y_t-1)
	(S_t-1_m) edge [connect] (Y_t-1)

	(S_t_1) edge [arc1] (Y_t)
	(S_t_2) edge [arc2] (Y_t)
	(S_t_m) edge [connect] (Y_t)

	(S_t+1_1) edge [arc1] (Y_t+1)
	(S_t+1_2) edge [arc2] (Y_t+1)
	(S_t+1_m) edge [connect] (Y_t+1)
	;
\end{tikzpicture}
			\subcaption{{\small The Factorial Hidden Markov Model.}}
			\label{fig:fhmm}
		\end{subfigure}
		\qquad 
		\begin{subfigure}[b]{0.45\textwidth}
			\begin{tikzpicture}[scale=0.8, font=\small, every node/.append style={transform shape}]
\tikzset{dummy/.style={}}
%for aligning the nodes with the original model
\tikzset{dummy_Y/.style={node distance = 13mm}}

\tikzstyle{var}=[circle, minimum size = 10.5mm, thick, draw =black!80, node distance = 5mm and 10mm]
\tikzstyle{connect}=[-latex, thick]
\tikzstyle{arc1}=[-latex, thick, bend left=40]
\tikzstyle{arc2}=[-latex, thick, bend left=30]
  \node[var] (S_t-1_1){$S_{t-1}^1$};
  \node[dummy, name=input1, left=of S_t-1_1] {};
  \node[var] (S_t_1) [right=of S_t-1_1] {$S_{t}^1$};
  \node[var] (S_t+1_1) [right=of S_t_1] {$S_{t+1}^1$};
  \node[dummy, name=output1, right=of S_t+1_1] {};
  \node[var] (S_t-1_2) [below=of S_t-1_1]{$S_{t-1}^2$};
  \node[dummy, name=input2, left=of S_t-1_2]{};
  \node[var] (S_t_2) [right=of S_t-1_2]{$S_{t}^2$};
  \node[var] (S_t+1_2) [right=of S_t_2]{$S_{t+1}^2$};
  \node[dummy, name=output2, right=of S_t+1_2] {};
  \node[var] (S_t-1_m) [below=of S_t-1_2]{$S_{t-1}^m$};
  \node[dummy, name=inputm, left=of S_t-1_m]{};
  \node[var] (S_t_m) [right=of S_t-1_m]{$S_{t}^m$};
  \node[var] (S_t+1_m) [right=of S_t_m]{$S_{t+1}^m$};
  \node[dummy, name=outputm, right=of S_t+1_m] {};
%dummy Y node for aligning with original model
  \node[dummy_Y, name=Y, below=of S_t-1_m]{};
  
 \path (input1) edge [connect] (S_t-1_1)
 	(S_t-1_1) edge [connect] (S_t_1)
        (S_t_1) edge [connect] (S_t+1_1)
        (S_t+1_1) edge [connect] (output1)        
        (input2) edge [connect] (S_t-1_2)
	(S_t-1_2) edge [connect] (S_t_2)
        (S_t_2) edge [connect] (S_t+1_2)
	(S_t+1_2) edge [connect] (output2)		
	(inputm) edge [connect] (S_t-1_m)
        (S_t-1_m) edge [connect] (S_t_m)
        (S_t_m) edge [connect] (S_t+1_m)
        (S_t+1_m) edge [connect] (outputm)
	;
\end{tikzpicture}
			\subcaption{{\small Structured variational approximation model}}
			\label{fig:variational}
		\end{subfigure}
\caption{{\small Factorial Hidden Markov Model and the variational approximation of it.}}
\label{fig:fhmm-models}
\end{figure}

The Factorial Hidden Markov Model (FHMM) \cite{factorial-hmms}\ is a bayesian graphical model in which a sequence of observed variables is generated from a latent sequence of discrete-valued vectors, as shown in Figure \ref{fig:fhmm}.  At a given time step $t \in \{1, \ldots, T\}$, the latent state $S_t$ is factored into $M$ dimensions, where each latent factor $S_t^m$ ($m \in \{1, \ldots, M\}$) is a vector of $K_m$ boolean variables with the constraint that exactly one of the boolean variables has value 1 at any time. We refer to the variable $S^m_{t,k}$ that has value 1 as the \emph{state} of layer $m$ at time step $t$.  In general the model could allow a different number of states $K_m$ for each layer $m$, but in our experiments we use a single value $K$ for all layers.  The observed variables $Y_t$ are also discrete variables, taking on $V$ possible values, where $V$ is the size of the vocabulary of words in the data.  

The model provides a joint distribution over these variables, which factors as follows:
\begin{align}
P(\{S_t,Y_t\}) & = P(S_1) P(Y_1|S_1) \prod_{t=2}^T {P(S_t|S_{t-1}) P(Y_t|S_t)}
\end{align}
where we refer to $P(S_1)$ as the \emph{initial distribution}, $P(Y_t|S_t)$ as the \emph{observation} distribution, and $P(S_t|S_{t-1})$ as the \emph{transition} distribution. 

The model assumes that states in one layer evolve independently of the other layers, and it uses log-linear models for the transition distribution of each layer.  This allows the transition distribution to factor as follows:
\begin{align}
P(S_t|S_{t-1}) 
& = \prod_{m=1}^M {P(S_t^m|S_{t-1}^m)} \\
& = \prod_{m=1}^M
	{
		\prod_{j=1}^K
			\frac
			{\prod_{k=1}^K 
			  { (\exp\theta_{mjk})^{S_{t-1,j}^m \cdot S_{t,k}^m}} 
			}
			{\sum_{k'=1}^K
			  (\exp\theta_{mjk'})^{ S_{t-1, j}^m }
			}
	}
\end{align}
where $\theta_{mjk}$ is the transition parameter of the model indexed by the layer, previous state and current state respectively. The initial distribution is defined the same way, except that we drop indicators for the previous time step and use parameters $\theta_{mk}$ rather than $\theta_{mjk}$.

\begin{comment}
Similarly, the initial probability distribution is defined as 
\begin{equation} 
P(S_1) = \prod_{m=1}^M {P(S_1^m)}
\end{equation}
\end{comment}

There are several possible functions to choose from for the observation distribution $P(Y_t|S_t)$. A table approach would need a matrix with $V \times K^M$ parameters, with exponential blow-up as we increase the number of layers.  Two standard ways of approximating this matrix of parameters are the noisy-or model and the log-linear model, and we choose log-linear distribution in our model.

The observations $Y_t$ are dependent on the states in all layers of $S_t$:
\begin{equation}
P(Y_t|S_t) = 
\frac { \prod_{m,k} 
    {(\exp\theta_{Y_tmk}) ^ {S_{t,k}^m}}
} 
{ \sum_Y 
    {\prod_{m,k} 
	{(\exp\theta_{Ymk})^{S_{t,k}^m}}
    }  
}
\end{equation}
where $\theta_{Ymk}$ is the observation parameter of the model indexed by the observed value, the latent layer, and the state of the latent layer respectively.

The number of parameters for the initial distribution is $M \times K$, for the transition distribution it is $M \times K^2$, and for the observation distribution it is $M \times K \times V$.

For finding representations using this model, we first estimate the parameters of the model by training it on unlabeled data in unsupervised fashion. Later, we use these parameters for finding the latent states for the data and use them as the representations.

While we borrow the model structure from Ghahramani and Jordan \cite{factorial-hmms}, the main between their model and our model is in the observation distribution.  Crucially, their model works with real-valued observations, while our model works with discrete-valued observations like words.  As a result of the change to the observation distribution, we need to change the inference and learning procedures, although we continue to follow \cite{factorial-hmms} in using variational approximations, which we explain below.

\section{Variational Methods for Learning and Inference}
\label{sec:learningandinference}

We aim to learn our representations from unlabeled text.  The objective for our unsupervised parameter estimation procedure is to maximize marginal log likelihood:
\begin{align}
\mathcal{L}(\theta) & = \log P(\{Y_t\}) = \log \sum_{\{S_t\}} P(\{S_t, Y_t\})  
\end{align}
We can re-write the objective using an arbitrary new distribution $Q(\{S_t\})$, which we call the \emph{variational} distribution, and apply Jensen's inequality, as follows:
\begin{align}
\log \sum_{\{S_t\}} P(\{S_t, Y_t\}) & = \log \sum_{\{S_t\}} {Q(\{S_t\}) \frac {P(\{S_t, Y_t\})} {Q(\{S_t\})}} \\
& \geq \sum_{\{S_t\}} {Q(\{S_t\}) \log \frac {P(\{S_t, Y_t\})} {Q(\{S_t\})}} = F(Q, \theta)
\end{align}
$F(Q, \theta)$ is a lower bound on the objective function that becomes exact when $Q$ is equal to the posterior distribution $P(\{S_t\} | \{Y_t\})$.  We can re-write $F(Q, \theta)$ in two different ways:
\begin{align}
F(Q, \theta) & = \sum_{\{S_t\}} {Q(\{S_t\}) \log P(\{Y_t\})} + \sum_{\{S_t\}} {Q(\{S_t\}) \log \frac{P(\{S_t\} | \{Y_t\})} {Q(\{S_t\})} } \nonumber \\
\label{eqn:kl-divergence} & = \mathcal{L}(\theta) - KL(Q(\{S_t\}) || P(\{S_t\} | \{Y_t\})) \\
\textrm{and } \nonumber \\
\label{eqn:entropy-difference} F(Q, \theta) & = E_{\{S_t\}\sim Q}[\log P(\{S_t, Y_t\})] - E_{\{S_t\}\sim Q}[\log Q(\{S_t\})] 
\end{align}
We use the Expectation Maximization (EM) algorithm \cite{Dempster77maximumlikelihood}, a block coordinate-ascent algorithm, to learn the parameters $\theta$.  The E-step consists of maximizing $F$ with respect to $Q$ while keeping $\theta$ fixed to the current set of parameters, which is equivalent to minimizing the KL-divergence in Equation \ref{eqn:kl-divergence}.  The M-step consists of maximizing $F$ with respect to $\theta$ while keeping $Q$ fixed, which is equivalent to maximizing the first term in Equation \ref{eqn:entropy-difference}. The algorithm guarantees that these two steps converge to a (local) maximum.

E-step: 
\begin{align} 
Q^{t+1} = \argmax_{Q} F(Q, \theta^t) = \argmin_{Q} KL(Q(\{S_t\}) || P(\{S_t\} | \{Y_t\}, \theta^t)
\end{align}
M-step:
\begin{align} 
\theta^{t+1} = \argmax_{\theta} F(Q^{t+1}, \theta) = \argmax_{\theta} E_{\{S_t\}\sim{Q^{t+1}}}[\log P(\{S_t, Y_t\}| \theta)]
\end{align}

\subsection{E-step}
In the E-step, we can minimize the KL-divergence to zero by setting the $Q$ distribution equal to the posterior distribution $P(\{S_t\} | \{Y_t\})$ if we can compute the posterior exactly. The posterior for the model is given by
\begin{equation}
P(\{S_t\} | \{Y_t\}) = 
\frac{1}{Z} 
\prod_t 
  \frac { \prod_{m,k} 
	    {(\exp\theta_{Y_tmk}) ^ {S_{t,k}^m}}
	} 
	{ \sum_Y 
	    {\prod_{m,k} 
		{(\exp\theta_{Ymk})^{S_{t,k}^m}}
	    }  
	}
\\
\prod_{t,m,j}
	{\frac
    {\prod_k 
      { (\exp\theta_{mjk})^{S_{t-1,j}^m \cdot S_{t,k}^m}} 
    }
    {\sum_{k'}
      (\exp\theta_{mjk'})^{ S_{t-1, j}^m }
    }
  }
\end{equation}
where, $Z=\sum_{\{S_t\}}{P(\{S_t, Y_t\})}$.

However, in an FHMM,  the exact computation of this posterior is computationally intractable, even using a forward-backward algorithm. The hidden state variables at each time-step become dependent conditioned on the observed variable requiring the computation of $K^M$ expectations, which is infeasible to compute for large $M$. 

For this reason, we resort to a variational approximation of the posterior $P$ by distribution $Q$ with its own variational parameters. The graphical model for the variational distribution $Q$ is chosen such that most of the structure of the original model is preserved, but such that the inference becomes tractable.  We borrow the structured variational approximation model from Ghahramani and Jordan \cite{factorial-hmms} which is shown in \ref{fig:variational}. The variational model is essentially $M$ independent Markov chains, one for each layer of the FHMM.  Markov chains permit efficient \emph{maximum a posteriori} inference using the standard Viterbi algorithm and efficient computation of marginal distributions using the standard forward-backward algorithm. 

The full variational distribution is given by
\begin{equation}
Q(\{S_t\} | \varphi) = 
\frac{1}{Z_Q} 
\prod_{t,m,k} 
  \exp(\varphi_{tmk})^{S_{t,k}^m} 
\\
\prod_{t,m,j} 
  {\frac
    {\prod_k 
      { (\exp\varphi_{mjk})^{S_{t-1,j}^m \cdot S_{t,k}^m}} 
    }
    {\sum_{k'}
      (\exp\varphi_{mjk'})^{ S_{t-1, j}^m }
    }
  }
\end{equation}
where $\varphi$ are the variational parameters of the model. $\varphi_{tmk}$ indicate observation variational parameters, and $\varphi_{mjk}$ indicate transition variational parameters.  The transition variational distribution matches the FHMM transition distribution.  Notice that the observation parameters are indexed by time step, rather than by the actual observation at that time step.  Thus the variational distribution can be thought of as having each layer a fictitious observation, chosen so as to mimic the true posterior as closely as possible.

The variational parameters are optimized by minimizing the KL-divergence, which obtains its minimum when the initial and transition variational parameters are equal to the initial and transition parameters of the original model, i.e. $\forall m,j,k\ .\ \varphi_{mjk} = \theta_{mjk}$ and $\varphi_{mk} = \theta_{mk}$. In this subspace of the parameter space, the KL-divergence simplifies to
\begin{align}
KL(Q || P) = & E \left[ - \log Z_Q + \log Z + \sum_{t,m,k} { S_{t,k}^m (\varphi_{tmk} - \theta_{Y_tmk})  }	\right] \nonumber \\
			 & + E \left[ \sum_t { \log \sum_Y \prod_{m} { \exp  \left(\sum_k  {\theta_{Ymk} S_{t,k}^m}  \right) } } \right] \label{eq:kldiv}
\end{align}
where, the expectations are taken using the $Q(\{S_t\})$ distribution.

Unfortunately, Equation \ref{eq:kldiv} still does not permit efficient optimization of the $\varphi$ parameters because of the log-sum-exp expression in the final term. While we followed the variational approximation from the original FHMM model so far, we provide a new algorithm to handle the log-sum-exp term introduced due to the observation distribution in our model. To handle this term, we make use of a second variational bound, $\log x \leq \phi x - \log \phi -1 $ \cite{variational-methods,2005-nips-corrlated-topic-model}, where $\phi$ is a new variational parameter which can be varied to make the bound tight. This allows us to rewrite the final term in \ref{eq:kldiv} as follows:
\begin{align}
E
	\left[ 
		\sum_t
		{
			\log \sum_Y \prod_{m}
				{
					\exp 
					\left(\sum_k 
					  {\theta_{Ymk} S_{t,k}^m} 
					\right)
				}
		}		
	\right]
& \leq
\sum_t
{
	\phi_t
	\left\{
	  \sum_Y
	  {
	    E
	    \left[
	    \prod_{m}
		    {
			    \exp 
			    \left(\sum_k 
			      {\theta_{Ymk} S_{t,k}^m} 
			    \right)
		    }				  
	    \right]
	  }
	  - \log \phi_t
	  - 1
	\right\}
} \label{eq:logbound}
\\
& =
\sum_t
{
	\phi_t
	\left\{
	  \sum_Y
	  {
	    \prod_{m}
	      E
	      \left[
		    {
			    \exp 
			    \left(\sum_k 
			      {\theta_{Ymk} S_{t,k}^m} 
			    \right)
		    }				  
	      \right]
	  }
	  - \log \phi_t
	  - 1
	\right\}
} \label{eq:distributionfactorization}
\\
& =
\sum_t
{
	\phi_t
	\left\{
	  \sum_Y
	  {
	    \prod_{m}
	      \left[
		    \sum_k
		    {
			    \exp (\theta_{Ymk})
			    E\left[
			      S_{t,k}^m
			    \right]
		    }				  
	      \right]
	  }
	  - \log \phi_t
	  - 1
	\right\}
} \label{eq:momentgenerating}
\end{align}

In Eqn. \ref{eq:distributionfactorization}, we move the expectation inside the product over $M$ independent layers. Eqn. \ref{eq:momentgenerating} follows from the moment generating function for the multinomial random variable $S_{t}^m$. 

%We also introduce additional notation $\langle \cdot \rangle$ to denote expectation.

Making use of this bound in Eqn. \ref{eq:kldiv}, we get a new upper bound on the KL-divergence, which we denote by $\overline{KL}$. 
The variational parameters $\phi$ have a closed-form solution when we set the gradient of $\overline{KL}$ with respect to the parameters $\phi$ to zero. We get $\phi_t = {\left( \sum_Y {\prod_m { \sum_{k'} E\left[ S_{t,k'}^m \right] } \exp\theta_{Ymk'}} \right)}^{-1}$.
To optimize the variational parameters $\varphi$, we find the gradient to minimize $\overline{KL}$.
\begin{align}
 \frac 
    {\partial \overline{KL}}
    {\partial \varphi_{tmk}} 
 = 
 \frac
    {\partial \overline{KL}}
    {\partial E\left[ S_{t,k}^m \right]} 
 \frac
    {\partial E\left[ S_{t,k}^m \right]} 
    {\partial \varphi_{tmk}}
\end{align}
Setting the derivative with respect to $E\left[ S_{t,k}^m \right]$ to zero, we get
\begin{align}
 \varphi_{tmk} 
 = 
 \theta_{Y_tmk} 
 - 
 \phi_t \sum_Y 
	\left\{
	  \left(
	    \prod_{n \neq m} 
	      {\sum_{k'} {E\left[ S_{t,k'}^m \right]} \exp \theta_{Ymk'}} \label{variationalparamoptimization}  
	  \right)
	  \exp\theta_{Ymk}
	\right\}
\end{align}
To find the optimized value for $\varphi_{tmk}$, we initially set the values to random values. We then run the forward-backward algorithm algorithm in each markov chain of the variational model and find the expectations $E\left[ S_{t,k}^m \right]$ using these variational parameters. Then these expectations are used to find the new set of variational parameters using the above equation. These two steps are iterated until convergence.

Once the variational parameters have been optimized, the variational distribution is used in the forward-backward algorithm in each layer of $Q$ for computing the expectations $E\left[ S_{t,k}^m \right]$ and $E\left[ S_{t-1,j}^m, S_{t,k}^m\right]$ required to compute the expected sufficient statistics for the M-step.

\subsection{M-step}
In the M-step, the best set of parameters of the model are found by maximizing the objective function
\begin{align}
F = E[\log P(\{S_t, Y_t\})]
\end{align}

Initial and transition parameters are normalized to proper distribution like in a standard HMM using the expected sufficient statistics gathered in the E-step. For the observation parameters, no closed form solution exists and we resort to gradient descent. We again use the bound on the objective function to move the expectation inside the parameter of the $\log$. The gradient of the lower bound $\overline{F}$ of the objective is
\begin{align}
\frac {\partial \overline{F}} {\partial \theta_{Ymk}} 
= \sum_t \left[ E\left[ S_{t,k}^m \right] 1[Y = Y_t] 
- \phi_t \left\{ \prod_{n \neq m} \left(\sum_{k'} { E\left[ S_{t,k'}^n \right] \exp\theta_{Ynk'}}\right) \right\} E\left[ S_{t,k}^m \right] \exp{\theta_{Ymk}}\right] \label{gradient}
\end{align}
where, $\phi_t = {\left( \sum_Y {\prod_m { \sum_{k'} E\left[ S_{t,k'}^m \right] } \exp\theta_{Ymk'}} \right)}^{-1}$.

\subsection{Inference}

To compute the posterior $P(\{S_t\} | \{Y_t\})$, we find the variational distribution $Q$ that best approximates the posterior using the same procedure as in the E-step.  $Q$ is also used to find the \emph{maximum a posteriori} state sequence using Viterbi algorithm.

\subsection{Implementation Details}
\label{implementation}
Rather than using Equations \ref{variationalparamoptimization} and \ref{gradient} naively, we precomputed few values to serve as a lookup for reducing the runtime complexity. First, we precomputed $\phi_t$ and also stored for each $Y$ the product over all $m$ (requiring memory $O(V)$). To compute the sum over all $Y$s, we reused these values, which only vary by a factor involving the dot product involving the layer $m$ and $Y$. We then used these computations to finally update the parameter in time $O(MKVT)$. We used similar precomputation technique in the M-step making the expensive gradient computation runtime $O(MKVT)$. Note that we can clear the memory after processing the result for a token making the memory use independent of the number of tokens. However, the runtime is dependent on the number of tokens. We used online learning technique \cite{Liang:2009:OEU:1620754.1620843,Bottou08thetradeoffs} for learning our model on large corpus. We took a mini-batch of 1000 sentences per iteration and ran for 5 epochs. We found that using observation parameters from previous iterations of EM to initialize the variational parameters allowed the variational parameters to converge much more quickly than random initialization. The variational parameter usually converged under 10 iterations. We used an existing package of L-BFGS for optimizing the parameters in the M-step, running for a maximum of 100 iterations.

\section{Experiments}
\label{sec:experiments}
We ran the experiments on two sequence labeling tasks in domain adaptation setting: part-of-speech (PoS) tagging and noun-phrase (NP) chunking. In both experiments, the source domain is from the newswire text and the target domain is from the biomedical text. However, the setting for the domain adaptation for the two experiments are different. In PoS tagging experiment, we get access to the text from the test domain and can use it to train the representation learning model. In the chunking experiment, however, we get access to a domain related to the test domain but not to the text from the test domain itself. In both experiments, latent vectors from the FHMMs are used as representations in the supervised experiments.

\subsection{Experimental Setup}
In the PoS tagging experiment, the labeled training data is from sections 02-21 of the Penn Treebank containing 39,832 sentences which have manually labeled PoS tags. We used a standard test set previously used by Huang \etal\ \cite{smoothing-HMM}\ and Blitzer \etal\ \cite{domain-adaptation-unsupervised}\ which contains 561 MEDLINE sentences with 14554 tokens from the Penn BioIE project. 
In the NP chunking experiment, we used the CoNLL-2000 shared task training data by replacing all non-NP chunk tags with outside chunk tag, O. The dataset consist of sections 15-18 of the Penn Treebank containing 8932 sentences which are annotated with PoS tags and the chunk tags\cite{Sang00introductionto}. The test set consist of 198 manually annotated sentences \cite{smoothing-HMM}\ from the biomedical domain of Open American National Corpus (OANC).

%data
For representation learning, we used the unlabeled data from both domains: WSJ corpus and from the MEDLINE corpus. We did not include any data from the biomedical domain of the OANC. Then we used a preprocessing step defined in \cite{Liang05semi-supervisedlearning}\ to select a subset of good sentences for training the FHMM, resulting in 112,824 sentences. This was done to train our system in reasonable amount of time but still learning a good representation from the data from both the domains. This data size is smaller than what previous researchers have used and covers fewer number of words in the supervised text. However, we think that a good representation for such words will be found by making use of our model which predicts it depending on the whole context of the sentence in the inference time. The words appearing only once in the corpus were collapsed into a special *unk* symbol and numbers were collapsed into a special *num* symbol, resulting in a vocabulary size of 40,596. This vocabulary did not cover 2\% and 0.3\% of the tokens in the supervised PoS training data and testing data respectively and 3\% of the tokens in both the training and test data for the chunking task.

We ran FHMMs with different state spaces and compared their representation capacity in the supervised experiments. We also compared the global context dependent distributed representations from FHMMs against other systems using the same unlabeled data for learning the representations. HMM provides globally context-dependent representations for the words which are not distributed representations. Brown Clusters provided a fixed cluster independent of the context and are not distributed representations. Word embeddings from neural network models also provided a fixed representation per word but are distributed representations. We trained 80 states HMM and used the Viterbi decoded states as representations \cite{smoothing-HMM}\ and a Brown Clustering algorithm \cite{1992-brown-clustering}\ with 1000 clusters as representations. Following Turian \etal\ \cite{2010-acl-word-representations-turian}, we used the prefixes of length 4, 6, 10 and 20 in addition to the whole path for the Brown representation as the features. We also trained a neural network model described in Collobert and Weston \cite{2008-icml-collobert-weston}\ and used previous 10 words to predict the next word in the sequence. We learned 50-dimensional word embeddings using 100 units in the hidden layer. We make the comparison of the representation systems by using them in the supervised experiments as features. We used Conditional Random Fields (CRF) with the same features as defined in Huang \etal\ \cite{smoothing-HMM}, with addition of the context words and context representation dimension as features in a window of size 2. In the chunking dataset, the representation learning text we used is different but the test set is the same as the previous work \cite{smoothing-HMM}\ we compare against, who also used the test sentences in the representation learning system.

We use the standard evaluation metric of word error rate for PoS tagging and F1 score for NP-chunking, focusing on the words that never appeared (OOV) and that appeared at most two times (rare) in the supervised training corpus. To calculate the precision for the OOV and rare words, we define false positive as the phrases predicted by the model beginning with the OOV or rare word but actually are not phrases in the gold dataset.

\subsection{Results and Discussion}
We represent the baseline system trained using traditional features with no features from a representation learning system as \baseline. \fhmmr{m}{k}\ is our system using Viterbi decoded states as representations from a $M$ layer FHMM with $K$ states in each layer. \fhmmpostr{m}{k}\ is our system using posterior probability distribution over states as representations from a $M$ layer FHMM with $K$ states in each layer. \hmm\ is a system using Viterbi decoded states as representation from an HMM model with 80 states and \brown\ is a system using Brown clustering algorithm with 1000 clusters as representations. The 50-dimensional word embeddings learned using neural network model from Collobert and Weston \cite{2008-icml-collobert-weston}\ is represented as \embedding. All these systems were trained on the same unlabeled representation text. In PoS tagging experiment, we report the error on all tokens (14,554). \huanghmm\ is a system defined in \cite{smoothing-HMM}\ and \scl\ is a system defined in \cite{domain-adaptation-unsupervised}\ and report the error rate on the same test data as ours but with fewer number of tokens (9576 and approx 13,000 respectively).  Also, they used unlabeled corpus of larger size to train their representation learning system. This makes the direct comparison with their system difficult. Table \ref{table:pos-tagging-accuracies} shows the error rate of the supervised classifier using different representation learning systems in PoS tagging experiment. Table \ref{table:np-chunking-result} shows the Recall(R), Precision(P) and F1 score of the supervised classifier using different representation learning system. Previous system \huanghmm\ \cite{smoothing-HMM} included the text from the test domain itself for representation learning.

We found that increasing the state space by increasing the number of layers in the representation learning system generally improved the performance of the supervised classifier. We also found model \fhmmpostr{5}{10} performs better than the same state size model \fhmmr{5}{10}. We think this is because the soft probabilities over the hidden states give more information to the supervised classifier compared to the one-hot representation as in Viterbi decoded states. Compared to the baseline, our best performing system reduced the error by 3.8\% on the OOV words and by 3.2\% on the rare words in the PoS tagging experiment and increased the F1 score by 5\% on both the OOV and the rare words in the chunking experiment.

We present analysis on why the performance of the different representation learning systems differ. \brown\ suffers mainly for two reasons. First, it assigns a fixed cluster to a word irrespective of its different usage as meaning in different domains. For example, the word \emph{share} which appears most frequently as NN in the train domain appears only as a VBP in the test domain and the word \emph{signaling} which only appears as VBG in the training domain appears only as NN in the test domain. Second, it provides no representation for the words that did not appear in the training of representation learning system. Models like HMM and FHMM do not suffer from these limitations. They both can provide different representations for the same word with different meaning in two domains. They also give a representation to a word by looking at the context even if it is not seen by the representation learning system. We also experimented by giving the context clusters when using the Brown clusters as representations but it did not improve the performance. \embedding\ also suffers in the PoS tagging experiment because of the fixed representations provided to the words irrespective of its different usage as meaning in different domains. In the chunking experiment, the performance of the \hmm\ model is worse than the baseline. We found that most of the errors were on the phrases surrounding the words like \emph{and} and punctuations where the \hmm\ system got confused on whether to continue the phrase or not whereas \fhmmr{5}{10} is performing much better in such situations. The word \emph{and} in the training text of the chunking data is clustered into fairly small (5-10) number of cluster in an 80 state HMM however it is clustered into 400 different clusters in FHMM model, usually varying in only few dimensions. Although we have not done further analysis, we think the representation using FHMM is able to capture larger context information into the representation which helps the supervised system to make better prediction. The reason \huanghmm\ performed better than the \baseline\ and \hmm\ might be because it is able to see the test sentences during the representation learning phase. The best results for the chunking task are provided by the distributed representations from the FHMMs or the word embeddings. The results show that the distributed representations from the FHMMs which can provide different representations per word depending on the whole sentence context can give either superior or comparable performance to the other representations.

\begin{table}
\begin{center}
\begin{tabular}{lccc}
\toprule
Model 						& All \% error 	&OOV \% error 	& Rare \% error	\\
\midrule
\baseline 					& 10.5			& 32.2			& 28.0	\\
\fhmmr{1}{10} 					& 10.4			& 31.7			& 27.8	\\
\fhmmr{3}{10} 					& 10.2			& 30.6			& 26.6	\\
\fhmmr{5}{10} 					& \bf{9.7}		& 29.7			& 25.2	\\
\fhmmpostr{5}{10}	 			& \bf{9.7}		& \bf{28.4}		& \bf{24.8} \\
\hmm 						& 10.1      		& 30.7			& 26.4		\\
\brown 						& 9.9      		& 30.2			& 25.8		\\
\embedding					& 10.2			& 30.3			& 26.3	\\
\midrule
\huanghmm* 					& 9.5      		& 24.8			& -		\\
\scl*						& 11.1			& 28.0			& - 	\\
\bottomrule
\end{tabular}
\end{center}

\caption{{\small  Error rate for the supervised classifier for the PoS tagging experiment using different representations. The error decreases as the state space size of the FHMM increases. Our model with highest state space size \fhmmr{5}{10} performs better than existing system trained using the representations learned on the same unlabeled text that we used. Also, the model \fhmmpostr{5}{10} with posterior probabilities as representation performs better than the model \fhmmr{5}{10} with Viterbi decoded states as representations. The bottom two rows represent systems trained on larger unlabeled corpus and reporting error on the same test set but with fewer number of tokens than ours (see text for details).}}
\label{table:pos-tagging-accuracies}
\end{table}

\begin{table}
\begin{center}
\begin{tabular}{lccccccccc}
\toprule
 			& \multicolumn{3}{c}{All}	 & \multicolumn{3}{c}{OOV} 	& \multicolumn{3}{c}{Rare}	\\
Model			& R & P & F1			& R & P & F1			& R & P & F1 \\ 
\midrule
%\baseline		&86.92 &89.74 & 88.31		&77.84 &88.92 & 83.01		&79.55  &88.61& 83.83	\\
\baseline		&.87 &.90 & .88		&.78 &.89 & .83				& .80 &.89 & .84	\\
%\fhmmr{1}{10}		&86.71 &89.58 &88.12		&77.29 &88.01 &82.30		&78.41 &87.56 &82.73	\\
\fhmmr{1}{10}		&.87 &.90 &.88			&.77 &.88 &.82			&.78 &.88 &.83	\\
%\fhmmr{3}{10}		&87.73 &89.96 &88.83		&80.33 &90.06 &84.92		&81.59 &89.53 &85.37	\\
\fhmmr{3}{10}		&.88 &.90 &.89			&.80 &.90 &.85			&.82 &.90 &.85	\\
\fhmmr{5}{10}		&.90 &\bf{.91} &.90		&.83 &.90 &.86			&.83 &.91 &.87	\\
%\fhmmpostr{5}{10}	&89.55 &\bf{91.22} &90.38	&85.04 &\bf{90.83} &\bf{87.84}	&85.91 &\bf{91.97} &\bf{88.84}	\\
\fhmmpostr{5}{10}	&.90 &\bf{.91} &.90		&.85 &\bf{.91} &\bf{.88}	&\bf{.86} &\bf{.92} &\bf{.89} \\
%\hmm 			&86.63 &89.85 &88.21    		&77.29 &87.19 &81.94		&79.09 &87.44 &83.05		\\
\hmm 			&.87 &.90 &.88      		&.77 &.87 &.82			&.79 &.87 &.83		\\
%\brown 			&87.07 &90.37 &88.69      	&78.95 &89.34 &83.82		&80.23 &89.59 &84.65		\\
\brown 			&.87 &.90 &.89      		&.79 &.90 &.84			&.80 &.90 &.85		\\
%\embedding		&89.04 &89.50 & 89.27		&\bf{85.60}&89.83&87.66		&\bf{86.14} &89.81&87.94	\\
\embedding		&.89 &.90 & .89			&\bf{.86} & .90 & \bf{.88}			&.86 &.90 & .88	\\
\midrule
\huanghmm* 		&\bf{.91} &.90 &\bf{.91}      	&.80 &.89 &.84			&.82 &.89 &.85		\\
\bottomrule
\end{tabular}
\end{center}

\caption{{\small Recall(R), Precision(P) and F1 score for the supervised chunking experiment using different representations. Our model \fhmmpostr{5}{10} performs better than all other models. \huanghmm\ is the HMM model system from a previous work on the same test set but using a different unlabeled representation text than ours (see text for details).}}
\label{table:np-chunking-result}
\end{table}

\section{Conclusion and Future Work}
\label{sec:conclusion}

We have developed a learning mechanism for discrete-valued FHMMs involving a novel two-step variational approximation that can train on datasets of nearly three million tokens in reasonable time.  Using the latent vectors predicted by our best FHMM model as features, we have shown that a standard classifier for two NLP tasks, chunking and tagging, and improve significantly on new domains, especially on words that were never or rarely observed in labeled training data.

Significant work remains to make the FHMM more useful as a representation learning technique.  Most importantly, it needs to scale to larger datasets using a distributed implementation.  More work is also needed to fully compare the relative merits of using global context against using local context in a representation learning framework.  The same idea of using global context to predict features can also be extended to other tasks, including hierarchical segmentation tasks like parsing, or perhaps even to image processing tasks using different classes of models.  Finally, it remains an open question whether there are efficient and useful ways to combine the distinct benefits of joint prediction and deep architectures for representation learning.

\section*{Acknowledgments}
This research was supported in part by NSF grant IIS-1065397.

\bibliographystyle{plain}

\bibliography{kia,representations,domain-adaptation,structured-learning,preprocessing,learning}

\begin{thebibliography}{10}

\bibitem{2010-naacl-ddowney-perplexity-vs-accuracy}
Arun Ahuja and Doug Downey.
\newblock Improved extraction assessment through better language models.
\newblock In {\em Proceedings of the Annual Meeting of the North American
  Chapter of the Association of Computational Linguistics (NAACL-HLT)}, 2010.

\bibitem{2009-bengio-curriculum-learning}
Y.~Bengio, J.~Louradour, R.~Collobert, and J.~Weston.
\newblock Curriculum learning.
\newblock In {\em International Conference on Machine Learning (ICML)}, 2009.

\bibitem{2008-neural-net-lang-models}
Yoshua Bengio.
\newblock Neural net language models.
\newblock {\em Scholarpedia}, 3(1):3881, 2008.

\bibitem{2005-nips-corrlated-topic-model}
David Blei and John Lafferty.
\newblock Correlated topic models.
\newblock In Y.~Weiss, B.~Sch\"{o}lkopf, and J.~Platt, editors, {\em Advances
  in Neural Information Processing Systems 18}, pages 147--154. MIT Press,
  Cambridge, MA, 2006.

\bibitem{lda}
David~M. Blei, Andrew~Y. Ng, and Michael~I. Jordan.
\newblock Latent dirichlet allocation.
\newblock {\em Journal of Machine Learning Research}, 3:993--1022, January
  2003.

\bibitem{domain-adaptation-theory}
John Blitzer, Koby Crammer, Alex Kulesza, Fernando Pereira, and Jenn Wortman.
\newblock Learning bounds for domain adaptation.
\newblock In {\em Advances in Neural Information Processing Systems}, 2007.

\bibitem{domain-adaptation-unsupervised}
John Blitzer, Ryan McDonald, and Fernando Pereira.
\newblock Domain adaptation with structural correspondence learning.
\newblock In {\em EMNLP}, 2006.

\bibitem{Bottou08thetradeoffs}
L\'eon Bottou and Olivier Bousquet.
\newblock The tradeoffs of large scale learning.
\newblock In {\em IN: ADVANCES IN NEURAL INFORMATION PROCESSING SYSTEMS 20},
  pages 161--168, 2008.

\bibitem{1992-brown-clustering}
P.~F. Brown, P.~V. deSouza, R.~L. Mercer, V.~J.~D. Pietra, and J.~C. Lai.
\newblock Class-based n-gram models of natural language.
\newblock {\em Computational Linguistics}, pages 467--479, 1992.

\bibitem{2009-parsing-with-brown-clusters}
M.~Candito and B.~Crabb´e.
\newblock Improving generative statistical parsing with semi-supervised word
  clustering.
\newblock In {\em IWPT}, pages 138--141, 2009.

\bibitem{2008-icml-collobert-weston}
R.~Collobert and J.~Weston.
\newblock A unified architecture for natural language processing: {D}eep neural
  networks with multitask learning.
\newblock In {\em International Conference on Machine Learning (ICML)}, 2008.

\bibitem{collobert:2011b}
R.~Collobert, J.~Weston, L.~Bottou, M.~Karlen, K.~Kavukcuoglu, and P.~Kuksa.
\newblock Natural language processing (almost) from scratch.
\newblock {\em Journal of Machine Learning Research}, 12:2493--2537, 2011.

\bibitem{frustratingly-easy-domain-adaptation}
Hal {Daum\'e III}.
\newblock Frustratingly easy domain adaptation.
\newblock In {\em ACL}, 2007.

\bibitem{2010-danlp-daume-semi-sup-domain-adaptation}
Hal {Daum\'e III}, Abhishek Kumar, and Avishek Saha.
\newblock Frustratingly easy semi-supervised domain adaptation.
\newblock In {\em Proceedings of the ACL Workshop on Domain Adaptation
  (DANLP)}, 2010.

\bibitem{max-ent-domain-adaptation-three-model-combo}
Hal {Daum\'e III} and Daniel Marcu.
\newblock Domain adaptation for statistical classifiers.
\newblock {\em Journal of Artificial Intelligence Research}, 26, 2006.

\bibitem{latent-semantic-indexing}
S.~C. Deerwester, S.~T. Dumais, T.~K. Landauer, G.~W. Furnas, and R.~A.
  Harshman.
\newblock Indexing by latent semantic analysis.
\newblock {\em Journal of the American Society of Information Science},
  41(6):391--407, 1990.

\bibitem{Dempster77maximumlikelihood}
A.~P. Dempster, N.~M. Laird, and D.~B. Rubin.
\newblock Maximum likelihood from incomplete data via the em algorithm.
\newblock {\em JOURNAL OF THE ROYAL STATISTICAL SOCIETY, SERIES B},
  39(1):1--38, 1977.

\bibitem{dhillon_icml12_tscca}
Paramveer~S. Dhillon, Jordan Rodu, Dean~P. Foster, and Lyle~H. Ungar.
\newblock Two step cca: A new spectral method for estimating vector models of
  words.
\newblock In {\em Proceedings of the 29th International Conference on Machine
  learning}, ICML'12, 2012.

\bibitem{downey07realm}
Doug Downey, Stefan Schoenmackers, and Oren Etzioni.
\newblock Sparse information extraction: {U}nsupervised language models to the
  rescue.
\newblock In {\em ACL}, 2007.

\bibitem{2008-emnlp-multi-source-adaptation}
Mark Dredze and Koby Crammer.
\newblock Online methods for multi-domain learning and adaptation.
\newblock In {\em Proceedings of EMNLP}, pages 689--697, 2008.

\bibitem{2010-machine-learning-multi-source-domain-adaptation}
Mark Dredze, Alex Kulesza, and Koby Crammer.
\newblock Multi-domain learning by confidence weighted parameter combination.
\newblock {\em Machine Learning}, 79, 2010.

\bibitem{2009-naacl-finkel-domain-adaptation-ner}
Jenny~Rose Finkel and Christopher~D. Manning.
\newblock Hierarchical bayesian domain adaptation.
\newblock In {\em Proceedings of HLT-NAACL}, pages 602--610, 2009.

\bibitem{factorial-hmms}
Zoubin Ghahramani and Michael~I. Jordan.
\newblock Factorial hidden markov models.
\newblock {\em Machine Learning}, 29(2-3):245--273, 1997.

\bibitem{grave2013hidden}
Edouard Grave, Guillaume Obozinski, and Francis Bach.
\newblock Hidden markov tree models for semantic class induction.
\newblock In {\em Proceedings of the Seventeenth Conference on Computational
  Natural Language Learning}, pages 94--103, Sofia, Bulgaria, August 2013.
  Association for Computational Linguistics.

\bibitem{2010-jmlr-henderson-sigmoid-belief-net}
James Henderson and Ivan Titov.
\newblock Incremental sigmoid belief networks for grammar learning.
\newblock {\em Journal of Machine Learning Research}, 2010.

\bibitem{1997-self-organizing-maps-as-reps}
T.~Honkela.
\newblock Self-organizing maps of words for natural language processing
  applications.
\newblock In {\em Proceedings of the International ICSC Symposium on Soft
  Computing}, 1997.

\bibitem{2014-compling-huang-lvlm-rep-learning}
Fei Huang, Arun Ahuja, Doug Downey, Yi~Yang, Yuhong Guo, and Alexander Yates.
\newblock {Learning Representations for Weakly Supervised Natural Language
  Processing Tasks}.
\newblock {\em Computational Linguistics}, 40(1), 2014.

\bibitem{smoothing-HMM}
Fei Huang and Alexander Yates.
\newblock Distributional representations for handling sparsity in supervised
  sequence labeling.
\newblock In {\em Proceedings of the Annual Meeting of the Association for
  Computational Linguistics (ACL)}, 2009.

\bibitem{2010-danlp-lvlms-for-domain-adaptation}
Fei Huang and Alexander Yates.
\newblock Exploring representation-learning approaches to domain adaptation.
\newblock In {\em Proceedings of the ACL 2010 Workshop on Domain Adaptation for
  Natural Language Processing (DANLP)}, 2010.

\bibitem{huang-open-domain-srl}
Fei Huang and Alexander Yates.
\newblock Open-domain semantic role labeling by modeling word spans.
\newblock In {\em Proceedings of the Annual Meeting of the Association for
  Computational Linguistics (ACL)}, 2010.

\bibitem{factorial-hmms-backfitting}
Robert~A. Jacobs, Wenxin Jiang, and Martin~A. Tanner.
\newblock {Factorial Hidden Markov Models and the Generalized Backfitting
  Algorithm}.
\newblock {\em Neural Computation}, 14(10):2415--2437, 2002.

\bibitem{instance-weighting}
Jing Jiang and ChengXiang Zhai.
\newblock Instance weighting for domain adaptation in {NLP}.
\newblock In {\em ACL}, 2007.

\bibitem{variational-methods}
Michael~I. Jordan, Zoubin Ghahramani, Tommi~S. Jaakkola, and Lawrence~K. Saul.
\newblock An introduction to variational methods for graphical models.
\newblock {\em Machine Learning}, 37(2):183--233, 1999.

\bibitem{1998-kaski-dim-reduction-by-random-mapping}
S.~Kaski.
\newblock Dimensionality reduction by random mapping: {F}ast similarity
  computation for clustering.
\newblock In {\em IJCNN}, pages 413--418, 1998.

\bibitem{2008-koo-dep-parsing-with-word-clusters}
T.~Koo, X.~Carreras, and M.~Collins.
\newblock Simple semi-supervised dependency parsing.
\newblock In {\em Proceedings of the Annual Meeting of the Association of
  Computational Linguistics (ACL)}, pages 595--603, 2008.

\bibitem{Liang05semi-supervisedlearning}
Percy Liang.
\newblock Semi-supervised learning for natural language.
\newblock In {\em MASTER'S THESIS, MIT}, 2005.

\bibitem{Liang:2009:OEU:1620754.1620843}
Percy Liang and Dan Klein.
\newblock Online em for unsupervised models.
\newblock In {\em Proceedings of Human Language Technologies: The 2009 Annual
  Conference of the North American Chapter of the Association for Computational
  Linguistics}, NAACL '09, pages 611--619, Stroudsburg, PA, USA, 2009.
  Association for Computational Linguistics.

\bibitem{2009-acl-ner-with-word-clusters}
D.~Lin and X~Wu.
\newblock Phrase clustering for discriminative learning.
\newblock In {\em ACL-IJCNLP}, pages 1030--1038, 2009.

\bibitem{2009-nips-multi-source-domain-adaptation-theory}
Y.~Mansour, M.~Mohri, and A.~Rostamizadeh.
\newblock Domain adaptation with multiple sources.
\newblock In {\em Advances in Neural Information Processing Systems}, 2009.

\bibitem{1998-trigram-brown-clustering}
S.~Martin, J.~Liermann, and H.~Ney.
\newblock Algorithms for bigram and trigram word clustering.
\newblock {\em Speech Communication}, 24:19--37, 1998.

\bibitem{2009-nips-mnih-hierarchical-lang-model}
A.~Mnih and G.~E. Hinton.
\newblock A scalable hierarchical distributed language model.
\newblock In {\em Neural Information Processing Systems (NIPS)}, pages
  1081--1088, 2009.

\bibitem{1993-acl-pereira-distributional-clustering}
F.~Pereira, N.~Tishby, and L.~Lee.
\newblock Distributional clustering of {E}nglish words.
\newblock In {\em Proceedings of the Annual Meeting of the Association for
  Computational Linguistics (ACL)}, pages 183--190, 1993.

\bibitem{2005-random-indexing-for-word-reps}
M.~Sahlgren.
\newblock An introduction to random indexing.
\newblock In {\em Methods and Applications of Semantic Indexing Workshop at the
  7th International Conference on Terminology and Knowledge Engineering (TKE)},
  2005.

\bibitem{2006-sahlgren}
M.~Sahlgren.
\newblock {\em The word-space model: {U}sing distributional analysis to
  represent syntagmatic and paradigmatic relations between words in
  high-dimensional vector spaces}.
\newblock PhD thesis, Stockholm University, 2006.

\bibitem{saltonIR}
G.~Salton and M.J. McGill.
\newblock {\em Introduction to Modern Information Retrieval}.
\newblock McGraw-Hill, 1983.

\bibitem{Sang00introductionto}
Erik F. Tjong~Kim Sang, Sabine Buchholz, and Kim Sang.
\newblock Introduction to the conll-2000 shared task: Chunking, 2000.

\bibitem{socher-acl-2013}
Richard Socher, John Bauer, Christopher~D. Manning, and Andrew~Y. Ng.
\newblock {Parsing with Compositional Vector Grammars}.
\newblock In {\em Association for Computational Linguistics ACL)}, 2013.

\bibitem{2007-acl-titov-sigmoid-belief-net}
Ivan Titov and James Henderson.
\newblock Constituent parsing with incremental sigmoid belief networks.
\newblock In {\em Association for Computational Linguistics ACL}, 2007.

\bibitem{2010-acl-word-representations-turian}
Joseph Turian, Lev Ratinov, and Yoshua Bengio.
\newblock Word representations: {A} simple and general method for
  semi-supervised learning.
\newblock In {\em Proceedings of the Annual Meeting of the Association for
  Computational Linguistics (ACL)}, pages 384--394, 2010.

\bibitem{2010-jair-turney-pantel}
P.~D. Turney and P.~Pantel.
\newblock From frequency to meaning: Vector space models of semantics.
\newblock {\em Journal of Artificial Intelligence Research}, 37:141--188, 2010.

\bibitem{2007-ica-for-semantic-features}
J.~J. V\"ayrynen, T.~Honkela, and L.~Lindqvist.
\newblock Towards explicit semantic features using independent component
  analysis.
\newblock In {\em Proceedings of the Workshop Semantic Content Acquisition and
  Representation (SCAR)}, 2007.

\bibitem{semi-supervised-embedding-srl}
Jason Weston, Frederic Ratle, and Ronan Collobert.
\newblock Deep learning via semi-supervised embedding.
\newblock In {\em Proceedings of the 25th International Conference on Machine
  Learning}, 2008.

\bibitem{zhao-conll09}
Hai Zhao, Wenliang Chen, Chunyu Kit, and Guodong Zhou.
\newblock Multilingual dependency learning: A huge feature engineering method
  to semantic dependency parsing.
\newblock In {\em {CoNLL} 2009 Shared Task}, 2009.

\end{thebibliography}

\end{document}